\documentclass{article}

\usepackage{amsmath, amsthm, amssymb, hyperref, color, float, graphicx, quiver, mathrsfs, authblk, circuitikz}

\oddsidemargin 
\evensidemargin
\newtheorem{theorem}{Theorem}
\newtheorem{example}[theorem]{Example}

\renewcommand{\paragraph}[1]{ \medskip \noindent {\bf #1.}}

\newcommand{\RR}{\mathbb{R}}
\newcommand{\NN}{\mathbb{N}}

\newcommand{\PP}{\mathbb{P}}
\newcommand{\EE}{\mathbb{E}}

\newcommand{\Id}{\text{Id}}
\newcommand{\Succ}{\text{Succ}}
\newcommand{\Spike}{\text{Spike}}

\title{Action-Driven Processes for Continuous-Time Control}
\author{Ruimin He, Shaowei Lin}
\affil{Ukusan Pte Ltd, Singapore}
\date{\today}

\begin{document}

\maketitle

\begin{abstract} \noindent
At the heart of reinforcement learning are \emph{actions} -- decisions made in response to observations of the environment. Actions are equally fundamental in the modeling of stochastic processes, as they trigger discontinuous state transitions and enable the flow of information through large, complex systems. In this paper, we unify the perspectives of stochastic processes and reinforcement learning through \emph{action-driven processes}, and illustrate their application to spiking neural networks. Leveraging ideas from \emph{control-as-inference}, we show that minimizing the Kullback-Leibler divergence between a policy-driven true distribution and a reward-driven model distribution for a suitably defined action-driven process is equivalent to maximum entropy reinforcement learning.
\end{abstract}

\section{Introduction}

Modeling systems that exhibit both continuous and discontinuous state changes presents a significant challenge in machine learning. For instance, biological spiking networks feature the continuous decay of neuron potentials alongside discontinuous spikes, which cause abrupt increases in the potentials of neighboring downstream neurons. Designing appropriate objective functions and applying gradient methods that work with these discontinuities are among the difficulties of working  with such systems.

Traditionally, ordinary and partial differential equations (ODEs and PDEs) are used to model continuous state changes, while Markov decision processes (MDPs) are employed to capture discrete actions that drive environmental transitions. In this paper, we study Action-Driven Processes (ADPs), also known as generalized semi-Markov processes \cite{matthes1962theorie,glynn1989gsmp,yu2015hidden}, which unify both types of dynamics within a single framework.  

With continuous-time states and actions at the core of ADPs, a natural question is whether it is possible to learn optimal policies for action selection using traditional reinforcement learning methods. The \emph{control-as-inference} tutorial \cite{levine2018reinforcement} elegantly demonstrated that maximum entropy reinforcement learning can be formulated as minimizing the Kullback-Leibler (KL) divergence between (a) a true trajectory distribution generated by action-state transitions and the policy, and (b) a model trajectory distribution that depends on the reward function. The demonstration involves a graphical model with binary random variables $\mathcal{O}_t$, each indicating whether the associated action $A_t$ is optimal. In this paper, we show that these optimality variables are not necessary in continuous-time ADPs. In fact, the connection between RL and variational inference arises naturally by comparing models in which the actions have independent arrival rates proportional to $e^{r(A_n, S_{n-1})}$ for some reward function $r(A_n,S_{n-1})$, with true distributions in which the actions arrive at some fixed rate $\rho$ but the actions are selected according to the policy $\pi_\theta(A_n\vert S_{n-1})$.

In Section \ref{sec:preliminaries}, we introduce continuous-time stochastic processes, with a focus on Markov and semi-Markov processes. Section \ref{sec:adp} defines action-driven processes in two equivalent ways and discusses their relationship to Markov decision processes. In Section \ref{sec:rl}, we demonstrate that maximum entropy reinforcement learning can be interpreted as variational inference on a suitable model for ADPs. Section \ref{sec:conclusion} concludes and outlines future steps. Throughout, we use spiking networks as illustrative examples to clarify the concepts.

\section{Preliminaries} \label{sec:preliminaries}

In this section, we assume that the reader is familiar with statistical models but not necessarily with stochastic processes. We motivate and introduce the definitions of stochastic processes, point processes, counting processes, Markov processes, and semi-Markov processes.

\subsection{Stochastic Processes}

A stochastic process \cite{albeverio2017probabilistic} on a probability space $(\Omega, \mathcal{F},\PP)$, indexed by a set $\mathcal{T}$, is a family of random variables $X(t) : \Omega \rightarrow \mathcal{X}$ for $t\in\mathcal{T}$, where $(\mathcal{X}, \mathcal{C})$ is a measurable space. We focus on \emph{finite joint distributions} of the form
\[
\PP(\{\omega \in \Omega: X(t_1,\omega) \in A_1, \ldots, X(t_n,\omega) \in A_n\})
\]
for a finite set $\{t_1, \ldots, t_n\}$ of indices and measurable subsets $A_1, \ldots, A_n \subset \mathcal{X}$. We write the above distribution as
\[
\PP( X(t_1) \in A_1, \ldots, X(t_n) \in A_n).
\]
The process is \emph{continuous-time} if the index set $\mathcal{T}$ is $[0,\infty) \subset \RR$. If $\mathcal{T}$ is the set $\NN = \{0, 1, 2, \ldots\}$ of natural numbers, the process is \emph{discrete-time}.

A stochastic process over some index set $\mathcal{T} \subset \RR$ is \emph{time-homogeneous} if
\[
\PP(X(t') = x' \vert X(t) = x ) = \PP(X(s') = x' \vert X(s) = x )
\]
for all $x', x \in \mathcal{X}$ and $s,s',t, t' \in \mathcal{T}$ such that $t'-t = s'-s$. Otherwise, it is \emph{time-inhomogeneous}.

A stochastic process indexed over a totally-ordered set $\mathcal{T}$ is said to be \emph{Markov} if
\begin{align}
\label{eq:markov-property}
\PP(X(t_{n+1}) = x_{n+1} \vert X(t_{n}) = x_{n}, \ldots, X(t_{0}) = x_{0} ) = \PP(X(t_{n+1}) = x_{n+1} \vert X(t_{n}) = x_{n} )
\end{align}
for all $t_0 \leq \ldots \leq t_{n+1}$ in $\mathcal{T}$ and all states $x_0, \ldots, x_{n+1} \in \mathcal{X}$. See \cite{weber2017master} for more details on continuous-time Markov processes.

\begin{example}[Boltzmann Machines] 

In a Boltzmann machine\cite{hinton2007boltzmann}, N neurons $\{x_0, x_1,... x_{N-1}\}$ are fully connected to each other. The input to neuron $i$ is \[z_{i} = b_i + \sum_{j} s_j w_{ij}\] where $b_i$ is the bias of neuron $i$, $w_{ij}$ is the symmetric weight of the connection between neurons $i$ and $j$ and $s_j$ is $1$ if the neuron $j$ is active and $0$ otherwise. At any time step, neuron $i$ is activated with probability \[\PP(x_i(t_{n+1}) = 1) = \frac{1}{1+e^{-z_i}}.\] In other words, the transition probability depends only on the previous state, making the Boltzmann machine a discrete-time neural network that forms a Markov chain.
\end{example}

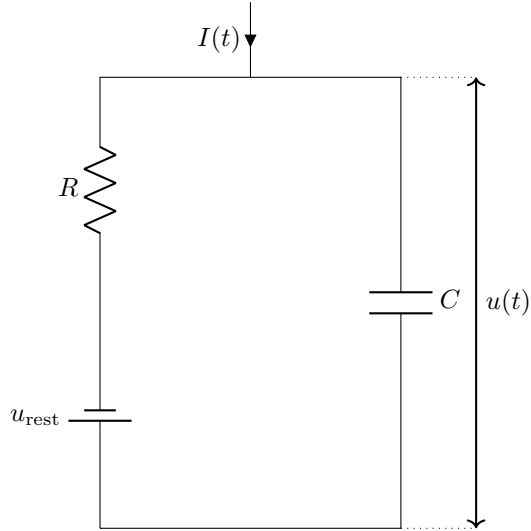
\begin{figure}[H]
\begin{center}
\begin{circuitikz}
  % Draw battery, resistor, and capacitor in series
  \draw
  (0,0) to[battery1, l=$u_\text{rest}$] (0,3)
  to[R, l=$R$] (0,6)
  to (4,6)
  to[C, l=$C$] (4,0)
  -- (0,0);

  \draw(2,7) to[short, i_=$I(t)$] (2,6);

  \draw[dotted] (4,6) -- ++(1,0);
  \draw[dotted] (4,0) -- ++(1,0);

  % Voltage arrow with label u(t)
  \draw[<->, thick] (5,0) -- (5,6)
    node[midway, right] {$u(t)$};  
\end{circuitikz}
\end{center}
\caption{A circuit model of an integrate-and-fire network}
\label{fig:rc_circuit}
\end{figure}

\begin{example}[Integrate-and-fire networks] \label{ex:iaf} Integrate-and-fire models \cite{gerstner2014neuronal} approximate a neuron as a leaky capacitor circuit (Figure \ref{fig:rc_circuit}), comprising a capacitor $C$ in parallel with a resistor $R$ and driven by a current $I$. The potential $u_i$ then evolves according to \[\tau_i \frac{du}{dt} = -(u(t) - u_\text{rest}) + RI(t)\] where $\tau_i = RC$ is the time constant and $u_\text{rest}$ is the resting potential of the circuit. In other words, the neuron's potential gradually decays towards $u_\text{rest}$ unless it is driven by incoming currents. The neuron fires once the potential crosses a threshold, $u_i(t) > u_\text{threshold}$, generating a current that feeds into other neurons. This exemplifies continuous-time single-neuron dynamics that forms a Markov chain, albeit without stochasticity.

\end{example}

\subsection{Point Processes}

Let $(\mathcal{X}, \mathcal{C})$ and $(\mathcal{Y}, \mathcal{D})$ be measurable spaces. A \emph{measure kernel} \cite{baccelli2024random} from $\mathcal{X}$ to $\mathcal{Y}$ is a map
\[
\xi:\mathcal{D} \times \mathcal{X} \rightarrow \bar{\RR}_{\geq 0}, \quad (D,x) \mapsto \xi(D|x)
\]
to the extended non-negative reals $\bar{\RR}_{\geq 0} := \RR_{\geq 0} \cup \{ \infty \}$, satisfying the following conditions:
\begin{itemize} 
\item For each $x\in \mathcal{X}$, $\xi(\,\cdot\,\vert x)$ is a measure on $\mathcal{D}$.
\item For each $D \in \mathcal{D}$, $\xi(D\vert \,\cdot\,)$ is a measurable function on $\mathcal{X}$.
\end{itemize}
Intuitively, $\xi(D|x)$ can be interpreted as a conditional measure on $\mathcal{Y}$ given $x$. If $\xi(\mathcal{Y}|x) = 1$ for all $x \in \mathcal{X}$, then $\xi$ is called a \emph{Markov kernel} or \emph{probability kernel}.

Often, we are interested in stochastic processes that describe \emph{arrivals} or other events that occur on $\mathcal{T}$. In particular, we will want to count these arrivals in measurable subsets $D \subset \mathcal{T}$. Let $\bar{\NN} := \NN \cup \{\infty\}$ be the extended natural numbers and let $(\mathcal{T}, \mathcal{D})$ be a measurable space. If $\xi$ is an \emph{$\bar{\NN}$-valued locally finite kernel} from $(\Omega, \mathcal{F})$ to $(\mathcal{T}, \mathcal{D})$, i.e.\ $\xi(\,\cdot\,\vert \omega)$ is a locally finite counting measure on $\mathcal{D}$ for all $\omega \in \Omega$, we define $\xi$ as a \emph{point process} on $\mathcal{T}$. 

Intuitively, for each measurable subset $D \subset \mathcal{T}$, the random variable $\xi(D) := \xi(D\vert \,\cdot\,):\Omega \rightarrow \bar{\NN}$ counts the number of events in $D$. Moreover, for each $\omega \in \Omega$, since $\xi(\,\cdot\,\vert \omega)$ is a locally finite counting measure, there is some locally finite multiset $T(\omega) \subset \mathcal{T}$ such that $\xi(D \vert \omega)$ is the number of points in $T(\omega) \cap D$ for each measurable $D \subset \mathcal{T}$. 

The \emph{mean measure} $\Lambda: \mathcal{D} \rightarrow \bar{\RR}_{\geq 0}$ of the point process $\xi$ is the expectation 
\[
\Lambda(D) := \EE[\xi(D)].
\]
If $\Lambda$ is absolutely continuous with respect to another measure $\mu \gg \Lambda$, the Radon-Nikodym derivative $\lambda = d\Lambda/d\mu$ is called the \emph{rate} or intensity, where
\[
\Lambda(D) = \int_D \lambda \, d\mu.
\]

If we have a family of point processes $\xi^{(\beta)}$, where $\beta > 0$ is a real-valued hyperparameter, such that
\[
\Lambda^{(\beta)}(D) := \EE[\xi^{(\beta)}(D)]\quad \text{and}\quad \Lambda^{(\beta)}(D) = \int_D \lambda^\beta \, d\mu,
\]
and $\lambda^\beta$ is the rate $\lambda$ of $\xi^{(1)}$ raised to the $\beta$ power, then we call $\beta$ the \emph{inverse temperature}. Its inverse $1/\beta$ is the \emph{temperature}. We will often consider the limiting behavior of $\xi^{(\beta)}$ as $\beta \rightarrow 0$ or $\beta \rightarrow \infty$. The latter is also called the \emph{zero temperature limit}.

\subsection{Counting Processes}

If $\xi$ is a point process on the continuous-time index set $\mathcal{T} = [0, \infty)$, we call the stochastic process $X(t) := \xi([0,t]) : \Omega \rightarrow \bar{\NN}$ is called a \emph{counting process}. Each sample $X(t,\omega)$ is piecewise constant and \emph{cadlag} (right continuous with left limits), so we define $X(t-) = \lim_{s \rightarrow t^{-}} X(s)$. Let $T(\omega)$ be the set of \emph{arrival times} $t$ where $X(t) \neq X(t-)$. The intervals $W(\omega)$ between consecutive arrival times are called \emph{wait times}. The counting process $X(t)$, arrival times $T(\omega)$, and wait times $W(\omega)$ are three equivalent ways of representing continuous-time point processes \cite[\S 7]{gerstner2014neuronal}.

Another useful perspective is to assign an \emph{action} $a(t)$ to each time $t$. An action $a: \mathcal{X}_a \rightarrow \mathcal{X}$ is a map from a subset $\mathcal{X}_a \subset \mathcal{X}$ of the state space to $\mathcal{X}$. Write the action of $a$ on a state $x$ as $ax$ rather than $a(x)$. Given a \emph{cadlag} function $x(t)$, we say that $a(t)$ \emph{generates} $x(t)$ if for all time $t$, $x(t) = a(t) x(t-)$. For counting processes $X(t)$, we consider the set $\{\Id, \Succ\}$ of actions, where $\Id$ is the \emph{identity} or \emph{trivial} action that maps each state $x \in \bar{\NN}$ to itself, and $\Succ$ is the \emph{successor} action that adds one to each state $x \in \bar{\NN}$. Let $A(t)$ be a stochastic process defined on the same probability space $(\Omega, \mathcal{F}, \PP)$ as $X(t)$. We say that $A(t)$ \emph{generates} $X(t)$ if $A(t, \omega)$ generates $X(t,\omega)$ for each $\omega \in \Omega$.

Let $X(s,t) := X((s,t])$ and $\Lambda(s,t) := \Lambda((s,t])$. The mean measure satisfies
\[
\Lambda(s, t) = \EE[X(s,t)].
\]
If the mean measure is absolutely continuous with respect to the Lebesgue measure $d\tau$ on $\mathcal{T}$, then the rate $\lambda(\tau)$ satisfies
\[
\Lambda(s, t) = \int_s^t \lambda(\tau)\, d\tau.
\]
This rate could depend on the current time $t$, the current state $X(t)$, or even past arrivals, especially if the point process is \emph{self-exciting} \cite{sahani2017describing}, for example, spiking neural networks.

We are interested in simple models where the point process is completely defined by the rate $\lambda(t)$. Recall that $N \sim \text{Pois}(\Lambda)$ follows the Poisson distribution with mean $\Lambda$ if 
\[
\PP(N = n) = \frac{\Lambda^n e^{-\Lambda}}{n!}, \quad n \in \NN.
\] 
A \emph{Poisson process} with rate $\lambda(t)$ and inverse temperature $\beta$ (for simplicity, assume $\beta=1$) is a point process with counts $X(t)$ where
\begin{itemize}
  \item for all intervals $(s,t]$, the random variable $X(s,t)$ follows $\text{Pois}(\int_s^t \lambda(\tau)^\beta\,d\tau)$;
  \item for all times $s<t\leq s'< t'$,  $X(s,t)$ and $X(s',t')$ are independent.
\end{itemize}
If $\lambda(t)$ is a constant, then the Poisson process is homogeneous; otherwise, it is inhomogeneous. 

To derive the distribution of the wait times, suppose that we start at time $s$ and want to find the probability that the first arrival occurs between time $t$ and $t+dt$ for some $dt > 0$. Then, for very small $dt$ and for all $t > s$, we have the conditional density
\begin{align*}
p(t|s)dt &= \PP(X(s,t)= 0)\, \PP(X(t,t+dt)=1) \\
&= \left(  e^{-\int_s^t \lambda(\tau)^\beta d\tau} \right) \left( \textstyle \int_t^{t+dt} \lambda(\tau)^\beta d\tau \,\, e^{-\int_t^{t+dt} \lambda(\tau)^\beta d\tau} \right) \\
&=  e^{-\int_s^t \lambda(\tau)^\beta d\tau} \,\lambda(t)^\beta\,dt.
\end{align*}
This formula reduces to the density of the exponential distribution with rate $\lambda^\beta$ when $\lambda(t) =\lambda$ is constant. Using the wait time density, we can express the density of paths $X(t)$ for $0\leq t \leq T$ corresponding to arrival times $0 < t_1 < \cdots < t_n < T$:
\begin{align*}
&p(t_1, \ldots, t_n) \,dt_1 \cdots dt_n \\ 
&= \left( e^{-\int_0^{t_1} \lambda(\tau)^\beta d\tau} \,\lambda(t_1)^\beta \,dt \right) \cdots \left( e^{-\int_{t_{n-1}}^{t_n} \lambda(\tau)^\beta d\tau} \,\lambda(t_n)^\beta \,dt \right) \left( e^{-\int_{t_n}^{T} \lambda(\tau)^\beta d\tau} \right) \\
&= e^{-\int_0^{T} \lambda(\tau)^\beta d\tau} \, \textstyle \prod_{i=1}^n \lambda(t_i)^\beta  \,dt_1 \cdots dt_n
\end{align*}
If $\lambda(t) =\lambda$ is constant, we can check that the path density integrates to $1$ over all $n$-simplices and over all path lengths $n \geq 0$:
\[
\textstyle \sum_{n=0}^\infty \int_{0 < t_1 < \cdots < t_n < T} p(t_1, \ldots, t_n) \,dt_1 \cdots dt_n =  \sum_{n=0}^\infty e^{- \lambda^\beta T} \lambda^{\beta n} \frac{T^n}{n!} =  e^{-\lambda^\beta  T}e^{\lambda^\beta  T} = 1.
\]

A useful generalization of the Poisson process involves relaxing the second independence condition above. We require that the rate $\lambda(t, t_a)$ depend only on the current time $t$ and the last arrival time $t_a$, but not on other arrivals or the current state $X(t)$. More precisely,
\begin{itemize}
  \item for all times $s < t \leq u \leq s' < t'$,  $X(s,t)$ and $X(s', t')$ are conditionally independent, given that $X$ has an arrival at $u$.
\end{itemize}
We refer to these as \emph{renewal processes} \cite{goodman2006introduction} because they reset at each arrival. Each arrival may correspond to a reset of a spiking neuron after it fires, or to the replacement of a faulty machine after it fails. If the rate depends only on the time $t-t_a$ since the last arrival, then we say that the renewal process is homogeneous; otherwise, it is inhomogeneous. In homogeneous renewal processes, the wait times are independent and identically distributed (i.i.d.).

\subsection{Discrete-Time Approximation}

To simulate a counting process with time-dependent rate $\lambda(t)$ in discrete time, we divide the time interval $[0, T]$ into $N$ intervals of length $\delta = T/N$ each. Since the number of arrivals in $(t, t+\delta]$ follows the Poisson distribution with mean $\int_t^{t+\delta}\lambda(\tau)^\beta d\tau$, we have 
\begin{align*}
\PP(X(t,t+\delta) = 0) &= e^{-\int_t^{t+\delta}\lambda(\tau)^\beta d\tau} \\
\PP(X(t,t+\delta) = 1) &= \textstyle \int_t^{t+\delta}\lambda(\tau)^\beta d\tau e^{-\int_t^{t+\delta}\lambda(\tau)^\beta d\tau} \\
&\vdots \\
\PP(X(t,t+\delta) = n) &= \textstyle \frac{1}{n!} \left(\int_t^{t+\delta}\lambda(\tau)^\beta d\tau \right)^n e^{-\int_t^{t+\delta}\lambda(\tau)^\beta d\tau} 
\end{align*}
For large $N$ and small $\delta$, if we have $\lambda(t') \approx \lambda(t)$ for all $t' \in (t, t+\delta]$, then
\begin{align} 
\notag \PP(X(t,t+\delta) = 0) &\approx e^{-\delta \lambda(t)^\beta } \\
\label{eq:discrete-time-approx} \PP(X(t,t+\delta) = 1) &\approx \delta \lambda(t)^\beta \,e^{-\delta \lambda(t)^\beta } \\
\notag &\vdots \\
\notag \PP(X(t,t+\delta) = n) &\approx \textstyle \frac{1}{n!} \left(\delta \lambda(t)^\beta \right)^n \,e^{-\delta \lambda(t)^\beta } 
\end{align}
If $\delta$ is sufficiently small, we can further ignore the cases where $X(t, t+\delta) \geq 2$ and assume that the variable $X(t, t+\delta)$ is binary.

For spiking neurons, the cell potential and the spike rate $\lambda(t)^\beta$ reset after each spike. The assumption that $\lambda(t') \approx \lambda(t)$ for all $t'$ in the interval $(t, t+\delta]$ does not generally hold throughout the process; however it remains valid for all $t'$ prior to the first arrival. The following approximation will be appropriate for the resulting renewal process. 
\begin{align} 
  \notag \PP(X(t,t+\delta) = 0) &\approx e^{-\delta \lambda(t)^\beta } \\ 
  \label{eq:discrete-time-approx-binary} \PP(X(t,t+\delta) \geq 1) &\approx 1- e^{-\delta \lambda(t)^\beta } 
\end{align}
For sufficiently small $\delta$, we will again have $\PP(X(t,t+\delta) \geq 1) \approx \delta \lambda(t)^\beta \,e^{-\delta \lambda(t)^\beta }$. The inequality $\geq 1$ here can be changed to an equality if we know that subsequent arrivals are highly unlikely after the first arrival.

In the zero temperature limit $\beta \rightarrow \infty$ and for fixed $\delta$, we see that 
\begin{align*}
  \PP(X(t,t+\delta) = 0) &\approx e^{-\delta \lambda(t)^\beta } \rightarrow 0 \quad \text{if} \log \lambda(t) > 0, \\
  \PP(X(t,t+\delta) = 0) &\approx e^{-\delta \lambda(t)^\beta } \rightarrow 1 \quad \text{if} \log \lambda(t) < 0. 
\end{align*}
Therefore, an arrival occurs almost surely if $\log \lambda(t) > 0$ but almost never if $\log \lambda(t) < 0$, just as with biological neurons, which fire if and only if a threshold is crossed.

\subsection{Semi-Markov Processes}

Previously, we studied many examples of stochastic processes in which the notion of an arrival is well-defined. Typically, arrivals correspond to discontinuities in a path $X(t)$ that is c\`adl\`ag. Arrivals could also be moments when the path changes state or exits some set, or when actions are chosen and executed. 

Recall that for homogeneous renewal processes, the wait times are identically distributed. We now look at processes where the distribution of the wait times depend on the state at the last arrival. Such processes are called \emph{semi-Markov processes}. Specifically, let $t_i$ denote the time of the $i$-th arrival, $w_{i} = t_{i}-t_{i-1}$ be the $i$-th wait time, and $t_0 = 0$. A semi-Markov process is one that satisfies, for all $n \geq 0$, time $t \geq 0$, and states $x_0, \ldots, x_{n+1}$,
\begin{align}
\notag &\PP(w_{n+1} \leq t, X(t_{n+1}) = x_{n+1} \vert X(t_n)=x_n, \ldots, X(t_0) = x_0) \\
\label{eq:semimarkov-property}&= \PP(w_{n+1} \leq t, X(t_{n+1}) = x_{n+1} \vert X(t_n)=x_n) \\
\notag &= \PP(w_{n+1} \leq t \vert X(t_n)=x_n) \,\PP( X(t_{n+1}) = x_{n+1} \vert X(t_n)=x_n).
\end{align}
where the next state $X(t_{n+1})$ is independent of the wait time $w_{n+1}$. Compare this equation to the Markov property \eqref{eq:markov-property}, which states that for all $n \geq 0$, times $0 \leq t_0 \leq \ldots \leq t_{n+1}$ and states $x_0, \ldots, x_{n+1}$, we have
\[
\PP(X(t_{n+1}) = x_{n+1} \vert X(t_{n}) = x_{n}, \ldots, X(t_{0}) = x_{0} ) = \PP(X(t_{n+1}) = x_{n+1} \vert X(t_{n}) = x_{n} )
\]
We see that for the semi-Markov property, the times $t_0, \ldots, t_{n+1}$ are only allowed to be arrival times. Hence, a process with arrivals that is Markov will also be semi-Markov. A \emph{discrete-time semi-Markov process} is a discrete-time stochastic process that satisfies the semi-Markov property \eqref{eq:semimarkov-property} where the times $t_0, t_1, \ldots$ and $t$ are non-negative integers.

If the process state remains constant between arrivals, i.e.\ $X(t) = X(t_n)$ for all $t_n \leq t < t_{n+1}$, then it is called a \emph{stepped} semi-Markov process. They were the first kind of semi-Markov processes to be studied \cite{levy1954processus,smith1955regenerative}. In general, the state $X(t)$ can vary between arrivals; here, we may call them \emph{continuous semi-Markov processes} \cite{harlamov2013continuous} to distinguish them from their stepped cousins. The states $X(t)$ between arrivals are often used to track inhomogeneous wait-time distributions. By extracting both the arrival times and corresponding states, $(t_0, x_0), (t_1, x_1), \ldots$, with $x_n = X(t_n)$, we obtain the \emph{embedded stepped semi-Markov process}, also known as the \emph{embedded Markov renewal process}. If we instead extract just the arrival states, $x_0, x_1, \ldots$, we recover the \emph{embedded discrete-time Markov process}.

\section{Action-Driven Processes} \label{sec:adp}

% Discrete-Time Semi-Markov Process
% https://dokumen.pub/hidden-semi-markov-models-theory-algorithms-and-applications-1nbsped-0128027673-978-0-12-802767-7.html
% https://apps.dtic.mil/sti/tr/pdf/ADA209264.pdf
% https://www.columbia.edu/~ww2040/ContinuityGSMP1980.pdf
% https://www.math.drexel.edu/~cmode/Presentations/presentation-semi-Markov-pdf 

Having established the preliminary foundations-namely, stochastic processes with arrivals-we have focused on the underlying \emph{states} and the transitions between them that occur at each arrival. We now shift our focus to \emph{actions}, assuming that transitions are triggered by a finite set of actions.

\subsection{Independent Action Arrivals}

At each current state, actions are generated by independent stochastic processes. The transition to the next state depends on the current state, the first arrival, and the time elapsed since the last action. In this article, we refer to them \emph{action-driven processes}, although in the literature they are more commonly known as \emph{generalized semi-Markov processes} \cite{matthes1962theorie,glynn1989gsmp,yu2015hidden}. Semi-Markov processes are a special case in which the transitions are triggered by a single non-trivial action. For example, the action set of a counting process is $\{\Id, \Succ\}$, where only $\Succ$ is non-trivial. Another example of an action-driven process is a spiking network with $n$ neurons, where the action set is $\{\Id, \Spike_1, \dots, \Spike_n\}$, and each $\Spike_i$ represents the spiking action of neuron $i$.
 
Specifically, given arrival times $t_i$ and wait times $w_i$, an action-driven process satisfies
\begin{align*}
&\PP(w_{n+1} \leq t, X(t_{n+1}) = x_{n+1} \vert X(t_n)=x_n, \ldots, X(t_0) = x_0) \\
&= \PP(w_{n+1} \leq t, X(t_{n+1}) = x_{n+1} \vert X(t_n)=x_n) \\
&= \textstyle \int_0^t \, \PP(\tau \leq w_{n+1} \leq \tau+d\tau, X(t_{n+1}) = x_{n+1} \vert X(t_n)=x_n) =: \textstyle \int_0^t \,\lambda^{(\beta)}_{x_n x_{n+1}}(\tau)\, d\tau
\end{align*}
for all $n \geq 0$, time $t \geq 0$ and states $x_0, \ldots, x_{n+1}$, where the \emph{transition rate} $\lambda^{(\beta)}_{xy}(t)$ from state $x$ to state $y$ at wait time $t$ expands to 
\begin{align*}
  \lambda^{(\beta)}_{xy}(t) \,dt &= \textstyle \sum_a \PP(t\leq w_{n+1} \leq t+dt, A(t_{n+1}) = a,  X(t_{n+1}) = y  \vert X(t_n)=x) \\
  &= \textstyle \sum_a  p_{xay}^{(\beta)}(t) \, \PP(t\leq w_{n+1} \leq t+dt, A(t_{n+1}) = a \vert X(t_n)=x), \\
  p_{xay}(t) &:= \PP( X(t_{n+1}) = y \vert w_{n+1}=t, A(t_{n+1}) = a, X(t_n)=x).
\end{align*}
Note that the \emph{transition probabilities} $p^{(\beta)}_{xay}(t)$ may depend on the wait time $t$. In the case of inverse temperatures $\beta \neq 1$, we define the distribution
$$
p^{(\beta)}_{xay}(t) = \frac{p_{xay}(t)^\beta}{\sum_y p_{xay}(t)^\beta}
$$
where $p_{xay}(t)$ are the transition probabilities at $\beta=1$. To compute the wait time densities $\PP(t\leq w_{n+1} \leq t+dt, A(t_{n+1}) = a \vert X(t_n)=x)$ we use
\begin{align}
  &\PP(t \leq w_{n+1} \leq t+dt, A(t_n+t) = a  \vert X(t_n)=x) \notag \\
  &= \textstyle \prod_{a'}\PP(\,\text{waiting till time } t\, \vert A(t_n+t) = a', X(t_n)=x) \,\cdot \notag \\
  & \qquad  \PP(\,\text{arrival between wait time } t \text{ and } t+dt \,\vert A(t_n+t) = a, X(t_n)=x)  \notag \\
  &= \textstyle \prod_{a'}e^{-\int_{0}^{t} \lambda_{xa'}(\tau)^\beta d\tau} \, \lambda_{xa}(t)^\beta\, dt \notag \\
  &= \textstyle e^{-\int_{0}^{t} \lambda^{(\beta)}_{x}(\tau) d\tau} \, \lambda_{xa}(t)^\beta\, dt \label{eq:independent-actions}
\end{align}
where the \emph{action rate} $\lambda_{xa}(t)^\beta$ is the arrival rate of action $a$ after wait time $t$ given the current state $x$, and the rate of \emph{any} arrival is given by the sum
$$
\lambda^{(\beta)}_x(t) = \sum_{a'} \lambda_{xa'}(t)^\beta.
$$

\begin{example}[Spiking network]
In the integrate-and-fire model (Example \ref{ex:iaf}), a neuron fires when the potential $u_i(t)$ crosses a threshold. This model can be modified such that the transition becomes stochastic rather than deterministic. For example, spikes could be modeled as a Poisson process, with the spiking rate $\lambda$ given by a monotonically increasing function of the potential $u_i(t)$. Once the spike occurs, the potentials of other neurons increase deterministically according to the strengths of their connections with the spiking neuron, i.e.,

\[\frac{du_{i}(t)}{dt} = -\frac{u_{i}(t)}{\tau}\]
between spikes, and when a neuron $j \neq i$ spikes, the potential jumps by
\[u_i \leftarrow u_i + \omega_{ij}.\]
The spikes are the arrivals of the process: each arrival changes the state by this jump, and between arrivals the potentials follow the deterministic decay.

\end{example}

\subsection{Embedded Stepped Processes}
\label{sec:stepped-processes}

By extracting the arrival times, actions and states of an action-driven process, 
\[(t_0, a_0, x_0), (t_1, a_1, x_1), \ldots, \quad \text{where each }a_n = A(t_n),\, x_n = X(t_n), w_n = t_n - t_{n-1}\]
we obtain the embedded \emph{stepped action-driven process}. From the previous section, we saw that the dynamics of this stepped process is completely determined by the action rates $\lambda_{xa}(t)$ and transition probabilities $p_{xay}(t)$. For the sake of generality, we allow trivial actions and the resulting transitions from a state $x$ back to itself, assuming that only finitely many of such actions are allowed in between non-trivial actions. Thus, different embedded stepped processes can be extracted from the same action-driven process.
% https://q.uiver.app/#q=WzAsMTMsWzIsMiwieF97bi0xfSJdLFszLDEsIndfblxcXFxhX24iXSxbNCwyLCJ4X3tufSJdLFs1LDEsIndfe24rMX1cXFxcYV97bisxfSJdLFswLDIsIlxcY2RvdHMiXSxbNiwyLCJcXGNkb3RzIl0sWzEsMl0sWzEsMSwid197bi0xfVxcXFxhX3tuLTF9Il0sWzMsMCwiXFxsYW1iZGFfe3hhfSh0KSJdLFs0LDEsInBfe3hheX0odCkiXSxbMSwwLCJcXGxhbWJkYV97eGF9KHQpIl0sWzUsMCwiXFxsYW1iZGFfe3hhfSh0KSJdLFsyLDEsInBfe3hheX0odCkiXSxbMCwxLCIiLDIseyJvZmZzZXQiOi0xfV0sWzAsMl0sWzQsMF0sWzIsNV0sWzIsMywiIiwwLHsib2Zmc2V0IjotMX1dLFsxLDJdLFszLDVdLFs3LDBdLFs0LDcsIiIsMSx7Im9mZnNldCI6LTF9XSxbOCwxLCIiLDAseyJzdHlsZSI6eyJib2R5Ijp7Im5hbWUiOiJkYXNoZWQifX19XSxbOSwyLCIiLDAseyJzdHlsZSI6eyJib2R5Ijp7Im5hbWUiOiJkYXNoZWQifX19XSxbMTAsNywiIiwwLHsic3R5bGUiOnsiYm9keSI6eyJuYW1lIjoiZGFzaGVkIn19fV0sWzEyLDAsIiIsMCx7InN0eWxlIjp7ImJvZHkiOnsibmFtZSI6ImRhc2hlZCJ9fX1dLFsxMSwzLCIiLDAseyJzdHlsZSI6eyJib2R5Ijp7Im5hbWUiOiJkYXNoZWQifX19XV0=
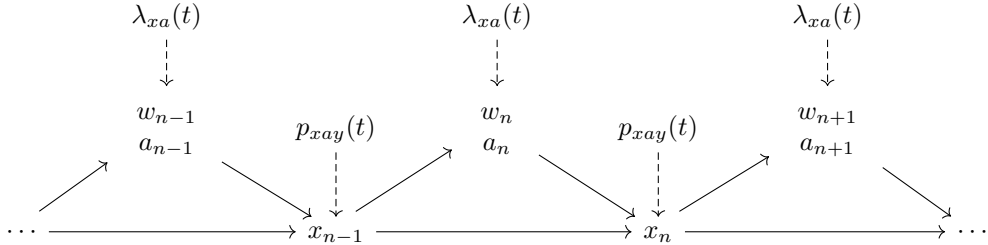
\[\begin{tikzcd}
	& {\lambda_{xa}(t)} && {\lambda_{xa}(t)} && {\lambda_{xa}(t)} \\
	& \begin{array}{c} w_{n-1}\\a_{n-1} \end{array} & {p_{xay}(t)} & \begin{array}{c} w_n\\a_n \end{array} & {p_{xay}(t)} & \begin{array}{c} w_{n+1}\\a_{n+1} \end{array} \\
	\cdots & {} & {x_{n-1}} && {x_{n}} && \cdots
	\arrow[dashed, from=1-2, to=2-2]
	\arrow[dashed, from=1-4, to=2-4]
	\arrow[dashed, from=1-6, to=2-6]
	\arrow[from=2-2, to=3-3]
	\arrow[dashed, from=2-3, to=3-3]
	\arrow[from=2-4, to=3-5]
	\arrow[dashed, from=2-5, to=3-5]
	\arrow[from=2-6, to=3-7]
	\arrow[shift left, from=3-1, to=2-2]
	\arrow[from=3-1, to=3-3]
	\arrow[shift left, from=3-3, to=2-4]
	\arrow[from=3-3, to=3-5]
	\arrow[shift left, from=3-5, to=2-6]
	\arrow[from=3-5, to=3-7]
\end{tikzcd}\]

Explicitly, given a finite set $\mathcal{A}$ of actions, each wait time $w_{n}^{(a)}$ where $a \in \mathcal{A}$ is independently sampled from the inhomogeneous Poisson process with rate $\lambda_{xa}(w)$, conditioned on the state $x_{n-1}$ at the last arrival time $t_{n-1}$. Its CDF is given by
\[
F(w)= 1-  \exp (-\int_0^w \lambda_{xa}(t_{n-1}+u) \,du).
\]
The wait time $w_n$ until the next arrival is given by the minimum of the action-specific wait times $w_{n}^{(a)}, a \in \mathcal{A},$ and the corresponding action $a_n$ is the argument attaining this minimum.
\begin{align*}
w_n &= \min_a w_n^{(a)}\\
a_n &= \arg \min_a w_n^{(a)}
\end{align*}
Finally, given $w_n, a_n$, the next state $x_n$ is drawn from the distribution $p_{xay}(w)$ where $x=x_{n-1}$, $a = a_n$, $y = x_{n}$, and $w = w_n$. The conditional relationships described above are summarized in the graphical model below.
% https://q.uiver.app/#q=WzAsNixbMCwzLCJ4X3tuLTF9Il0sWzEsMiwid19uXnsoYScpfSJdLFs0LDMsInhfe259Il0sWzEsMCwid19uXnsoYSl9Il0sWzEsMSwiXFx2ZG90cyJdLFszLDIsIndfbiA9IFxcbWluX2Egd19uXnsoYSl9XFxcXGFfbiA9IFxcYXJnIFxcbWluX2Egd19uXnsoYSl9Il0sWzAsMSwiIiwyLHsib2Zmc2V0IjotMX1dLFswLDJdLFswLDNdLFszLDVdLFsxLDVdLFs1LDJdXQ==
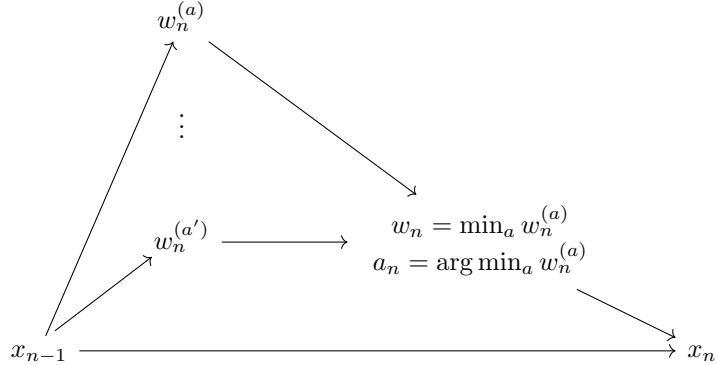
\[\begin{tikzcd}
	& {w_n^{(a)}} \\
	& \vdots \\
	& {w_n^{(a')}} && \begin{array}{c} w_n = \min_a w_n^{(a)}\\a_n = \arg \min_a w_n^{(a)} \end{array} \\
	{x_{n-1}} &&&& {x_{n}}
	\arrow[from=1-2, to=3-4]
	\arrow[from=3-2, to=3-4]
	\arrow[from=3-4, to=4-5]
	\arrow[from=4-1, to=1-2]
	\arrow[shift left, from=4-1, to=3-2]
	\arrow[from=4-1, to=4-5]
\end{tikzcd}\]

Suppose the embedded stepped process is currently in state $x$. Let $\ell_{xa}(t)$ denote the rate of action $a$ at time $t$ given $x$, where $a$ may include the trivial action. If action $a$ is not accessible at time $t$ given $x$, we set $\ell_{xa}(t) = 0$. Since the actions arrive independently, the rate of \emph{any} action occurring is $\ell_x(t) = \sum_{a} \ell_{xa}(t)$. Thus, the wait time can be sampled from a Poisson process with rate $\ell_x(t)$. Given that \emph{some} action occurs at time $t$, the probability of each action at time $t$ is given by
\[
p_{xa}(t) = \frac{\ell_{xa}(t)}{\sum_{a'} \ell_{xa'}(t)} = \frac{\ell_{xa}(t)}{\ell_x(t)}.
\]
For the trivial action $\Id$, if $\ell_{x,\Id}(t)$ is nonzero, then $p_{x,\Id}(t)$ is also non-zero. Overall, this stepped process simulates an action-driven process $X(t)$ with action rates $\lambda_{xa}(t) = \ell_{xa}(t)$ for all non-trivial $a$. By varying the rate $\ell_{x,\Id}(t)$, we obtain different embedded stepped processes that correspond to the same underlying action-driven process, with various action rates $\ell_x(t)$ and action probabilities $p_{xa}(t)$.

An inverse temperature parameter $\beta > 0$ can be introduced to the model by replacing rates $\ell_{xa}(t)$ with $\ell_{xa}(t)^\beta$. The action rates and action probabilities will then become
\[
\ell^{(\beta)}_x(t) = \sum_{a} \ell_{xa}(t)^\beta, \quad p^{(\beta)}_{xa}(t) = \frac{\ell_{xa}(t)^\beta}{\ell^{(\beta)}_x(t)}.
\]
In the zero temperature limit $\beta \rightarrow \infty$, the action probabilities end up converging to $1$ for the action $a$ with the highest rate $\ell_{xa}(t)$. However, the action rates diverge to infinity, implying that the arrivals occur instantaneously. Consequently, the limit can be interpreted as a deterministic stepped process, in which the rate-maximizing action is selected at each step and the actual wait times are disregarded.

A special case of the stepped process construction, known as \emph{uniformization}, consists of choosing a sufficiently large action rate $\lambda$ and setting $\ell^{(\beta)}_x(t) = \lambda$ for all states $x$. Let $X(t)$ be an action-driven process with action rates $\lambda_{xa}(t)^\beta$ for all non-trivial actions $a \neq \Id$. The action probabilities will be given by
\begin{align*}
p^{(\beta)}_{xa}(t) &= \frac{\lambda_{xa}(t)^\beta}{\lambda} \quad \text{for all }a \neq \Id,\\
p^{(\beta)}_{x,\Id}(t) &=  1- \sum_{a\neq \Id} \frac{\lambda_{xa}(t)^\beta}{\lambda} = 1 - \frac{\lambda^{(\beta)}_{x}(t)}{\lambda}.
\end{align*}
For these probabilities to be well-defined, we require $\lambda \geq \lambda^{(\beta)}_{x}(t)$ for all states $x$ and times $t$. Simulating this stepped process is relatively straightforward: action arrivals follow a Poisson process with rate $\lambda$ and can be generated by sampling wait times from an exponential distribution with rate $\lambda$. Alternatively, when sampling over a fixed time interval $[0, T]$, we first draw the number $N$ of arrivals from the Poisson distribution with mean $\lambda T$ 
\[
\PP(N=n) = \displaystyle e^{-\lambda T} \frac{(\lambda T)^n}{n!}.
\]
We then pick arrival times $t_1, \ldots, t_n$ independently and uniformly from the interval $[0,T]$. At each arrival, the next action is sampled according to the action probabilities $p^{(\beta)}_{xa}(t)$, which may include the trivial action. The subsequent state is then selected using the transition probabilities $p^{(\beta)}_{xay}(t)$. The equivalence between sampling by uniformization and sampling the original action-driven process follows from \cite[Thm~1]{van1992uniformization}.

As the uniformization rate $\lambda$ tends to infinity, we get more fine-grained embeddings of the original action-driven process, which can be used for approximating path integrals. In this limit, both stochastic and quantum path integrals exhibit well-behaved properties; see, for example, \cite[\S5]{gill2002foundations}.

\subsection{Action After Arrival}

In equation \ref{eq:independent-actions}, we assumed that the actions occur independently of each other, with each action having a wait time that is independent of the others. During the simulation, the action with the smallest sampled wait time is selected. We refer to this as the independent-action-arrivals (IAA) definition. 

An alternative definition of ADPs first samples the wait time to \emph{some} action, and then selects the action from a finite set according to a multinomial distribution. We refer to this as the action-after-arrival (AAA) definition. We will now show that this alternative definition is equivalent to the original IAA formulation.

As discussed in Section \ref{sec:stepped-processes}, an AAA definition can be derived from the IAA definition. We will now start from the AAA definition. Suppose that the wait time is generated by the inhomogeneous Poisson process with rate $\lambda_x(t)$ where $x$ is the state at the previous arrival. Suppose that an action is then selected from the multinomial distribution $p_{xa}(t)$, satisfying $\sum_a p_{xa}(t) = 1$. In the corresponding IAA model, the actions have independent arrival rates $\lambda_{xa}(t) = p_{xa}(t) \lambda_x(t)$. Then, as required, the arrival rate of \emph{any} action is 
\[
\lambda'_x(t) = \textstyle\sum_a \lambda_{xa}(t) = \textstyle\sum_a p_{xa}(t) \lambda_x(t) = (\textstyle\sum_a p_{xa}(t)) \lambda_x(t) = \lambda_x(t)
\]
and the probability of action $a$ arrived given that some action arrived is
\[
p'_{xa}(t) = \frac{\lambda_{xa}(t)}{\lambda_x(t)} = \frac{p_{xa}(t) \lambda_x(t)}{\lambda_x(t)} = p_{xa}(t).
\]

\subsection{Relationship to Markov Decision Processes}

A Markov Decision Process (MDP) models a system with discrete states $\{S_{0}, ..., S_{N-1}\}$, where at each discrete time step $t$, an agent selects an action $a \in A$. The resulting state transition matrix is described by a probability distribution that depends solely on the current state $s_{t}$ and action $a_{t}$. At every time step, the agent also receives a reward $R(s, a)$, and seeks to maximize the discounted sum of rewards over time. The objective in an MDP is typically to pick a policy that maximizes the expected cumulative reward over a potentially infinite horizon.

An embedded stepped action-driven process can be regarded as a special case of an MDP if the reward function is ignored. Its action-state transition matrix corresponds to that of the ADP, and its policy is defined by the two-step procedure for selecting an action: first sampling the wait time, then sampling the action conditional on the wait time. Conversely, the MDP can also be thought of as a special case of the ADP, in which the wait time is disregarded, and the arrival rates and action-state transitions are assumed to be independent of the wait time. 

When deciding whether to use ADPs or MDPs to model a complex system, several considerations are key. First, ADPs are more suited for modeling continuous-time systems with discontinuous state changes occurring at specific arrival times. Second, ADPs naturally model concurrent systems in which computation is driven by actions. Third, ADPs are advantageous for machine learning or reinforcement learning in such concurrent systems, because they abstract away the gradual state changes between arrivals - periods during which information does not flow between system components but which may still influence arrival rates.

\section{Reinforcement Learning} 
\label{sec:rl}

Let the state of the environment at the $n$-th arrival be $S_n$, and let the action chosen at that arrival be $A_n$. Consider a reinforcement learning (RL) problem with a reward function $r(A_n, S_{n-1})$ and a policy $\pi_\theta(A_n\vert S_{n-1})$ parametrized by $\theta$. The environment has an initial distribution $q(S_0)$ and its response to an action is described by the transition distribution $q(S_n\vert A_n, S_{n-1})$. Although these environmental distributions are unknown, we assume that we can sample from them. 

We now show how to represent this RL problem as an inference problem on continuous-time ADPs, using ideas from \emph{control-as-inference} \cite{levine2018reinforcement}.  Let $W_n$ denote the wait time between the $(n-1)$-th and $n$-th arrivals. We define two distributions - the true distribution $q$ and the model distribution $p$ - and consider the Kullback-Leibler (KL) divergence of $q$ from $p$. By construction, the true distribution $q$ depends on the policy $\pi_\theta(A_n\vert S_{n-1})$, whereas the model distribution $p$ is influenced by the reward $r(A_n, S_{n-1})$.

In the true distribution, we assume that the wait times are exponentially distributed with constant rate $\rho$. The action distribution $q(A_n \vert W_n, S_{n-1})$ is given by $\pi_\theta(A_n \vert S_{n-1})$, and is independent of the wait time. The trajectory density is given by
\begin{align*} 
  q(S_{0\ldots N}, W_{1\ldots N}, A_{1 \ldots N}) & := q(S_0) \textstyle \prod_{n=1}^N q(W_n \vert S_{n-1})  \,q(A_n \vert S_{n-1})\,q(S_n \vert A_n,  S_{n-1}) \, \\
  q(S_0) &\quad \text{true initial distribution} \\
  q(S_n|A_n, S_{n-1})  &\quad \text{true transition distribution} \\
  q(W_n\vert S_{n-1}) &:= e^{-\rho W_n} \rho\,dW_n \\
  q(A_n\vert S_{n-1}) &:= \pi_\theta(A_n|S_{n-1})
\end{align*}

In the model distribution, we set the environmental distributions to match the true distributions, with $p(S_0) = q(S_0)$ and $p(S_n|A_n, S_{n-1})=q(S_n|A_n, S_{n-1})$. For the actions, we assume that they have distinct arrival rates $\lambda(A_n, S_{n-1})$, which are influenced by the rewards $r(A_n, S_{n-1})$ via
\[
\lambda(A_n, S_{n-1}) = e^{r(A_n, S_{n-1})}.
\]
Let $\lambda(S_{n-1})$ be the sum $\sum_a \lambda(a, S_{n-1})$. Then, the trajectory density is given by
\begin{align*} 
  p(S_{0\ldots N}, W_{1\ldots N}, A_{1 \ldots N}) &:= p(S_0) \textstyle \prod_{n=1}^N p(W_n, A_n \vert S_{n-1})  \,p(S_n \vert A_n, S_{n-1}) \\
  p(S_0) &\quad \text{model initial distribution} \\
  p(S_n|A_n, S_{n-1})  &\quad \text{model transition distribution} \\
  p(W_n, A_n\vert S_{n-1}) &:= e^{-\lambda(S_{n-1})W_n} \lambda(A_n, S_{n-1})\, dW_n
\end{align*}

To find the best policy $\pi_\theta$, we shall minimize over $\theta$ the KL divergence of $q$ from $p$. 
\[
I_{q\Vert p}(S_{0\ldots N}, W_{1\ldots N}, A_{1 \ldots N}) = \int q(s,w,a) \log \frac{q(s,w,a)}{p(s,w,a)} \,ds\, dw\, da.
\]
Applying the chain rule of relative information $I(X,Y) = I(X|Y) + I(Y)$, we get 
\begin{align*}
& I_{q\Vert p}(S_{0\ldots N}, W_{1\ldots N}, A_{1 \ldots N})\\ 
&= I_{q\Vert p}(S_{0}) +  \textstyle \sum_{n=1}^N  I_{q\Vert p}(S_{n}, W_{n}, A_{n}\vert S_{n-1}, W_{n-1}, A_{n-1}) \\
&= I_{q\Vert p}(S_{0}) + \textstyle \sum_{n=1}^N I_{q\Vert p}(W_{n}, A_{n}\vert S_{n-1}) + I_{q\Vert p}(S_{n} \vert A_n, S_{n-1}) 
\end{align*}
Because $p(S_0)=q(S_0)$ and $p(S_{n} \vert A_n, S_{n-1}) = q(S_{n} \vert A_n, S_{n-1})$, the terms $I_{q\Vert p}(S_{0})$  and $I_{q\Vert p}(S_{n} \vert A_n, S_{n-1})$ vanish. Therefore,
\begin{align*}
& I_{q\Vert p}(S_{0\ldots N}, W_{1\ldots N}, A_{1 \ldots N})\\ 
&= \textstyle \sum_{n=1}^N I_{q\Vert p}(W_{n}, A_{n}\vert S_{n-1}) \\
&= - \textstyle \sum_{n=1}^N \EE_{q(W_n,A_n,S_{n-1})}[\log p(W_{n}, A_{n}\vert S_{n-1}) - \log q(W_{n}, A_{n}\vert S_{n-1})] \\
&= - \textstyle \sum_{n=1}^N\EE_{q(W_n,A_n,S_{n-1})}[r(A_n, S_{n-1}) -\lambda(S_{n-1}) W_n -\log \pi_\theta(A_n\vert S_{n-1})+\rho W_n - \log \rho] \\ 
&= \textstyle \sum_{n=1}^N \Big( -\EE_{q(W_n,A_n,S_{n-1})}[r(A_n, S_{n-1}) -\log \pi_\theta(A_n\vert S_{n-1})] \\
& \quad \quad +\EE_{q(S_{n-1})}[\lambda(S_{n-1})]\,\EE_{q(W_n)}[W_n] -\rho\, \EE_{q(W_n)}[W_n] + \log \rho \Big) \\ 
&= - \textstyle \sum_{n=1}^N\EE_{q(W_n,A_n,S_{n-1})}[r(A_n, S_{n-1}) -\log \pi_\theta(A_n\vert S_{n-1})] \\
&\quad \quad +\rho^{-1} \textstyle \sum_{n=1}^N \EE_{q(S_{n-1})}[\lambda(S_{n-1})] + N(\log \rho - 1),
\end{align*}
where we used that under $q$ the wait time $W_n$ is independent of $S_{n-1}$, with $\EE_{q(W_n)}[W_n] = \rho^{-1}$. The constant $N(\log \rho - 1)$ does not depend on $\theta$. The term $\rho^{-1} \sum_{n} \EE_{q(S_{n-1})}[\sum_a e^{r(a, S_{n-1})}]$ does depend on $\theta$, through the state distribution, and it penalizes policies that reach states where the total reward is high. This bias works against the RL objective, and it is what the large-$\rho$ condition suppresses: it is at most $N\rho^{-1} \sup_s \sum_a e^{r(a,s)}$, which is negligible when $\rho \gg \sup_s \sum_a e^{r(a,s)}$. So for sufficiently large $\rho$, minimizing the original objective $I_{q\Vert p}(S_{0\ldots N}, W_{1\ldots N}, A_{1 \ldots N})$ is approximately the same as maximizing
\begin{align*}
&\textstyle \sum_{n=1}^N\EE_{q(W_n,A_n,S_{n-1})}[r(A_n, S_{n-1}) -\log \pi_\theta(A_n\vert S_{n-1})] \\
&= \textstyle \sum_{n=1}^N\EE_{q(W_n,A_n,S_{n-1})}[r(A_n, S_{n-1})]  +\EE_{q(S_{n-1})}[ - \sum_{A_n} \pi_\theta(A_n\vert S_{n-1}) \log \pi_\theta(A_n\vert S_{n-1})] \\
&= \textstyle \sum_{n=1}^N\EE_{q(W_n,A_n,S_{n-1})}[r(A_n, S_{n-1})]  +\EE_{q(S_{n-1})}[ H( \pi_\theta(A_n\vert S_{n-1}))] 
\end{align*}
where the last summand is the expected conditional entropy. 

This yields the objective function commonly employed in \emph{maximum entropy reinforcement learning}. Note that the optimality variables described in the control-as-inference paper \cite{levine2018reinforcement} are not required here.

\section{Conclusion}
\label{sec:conclusion}

To summarize, we proposed an action-centric perspective of continuous-time models that incorporate both continuous and discontinuous changes of state. We presented two equivalent definitions of action-driven processes and their relationship to Markov decision processes. Finally, we demonstrated that maximum entropy reinforcement learning can be interpreted as variational inference on a simple class of ADPs.

In future work, we will model different kinds of spiking neural networks using ADPs, and to develop learning algorithms based on information-theoretic strategies derived from variational inference. We will also explore the category theoretic foundations of ADPs where the objects are states and the morphisms are actions of some symmetric monoidal category. In particular, we will look at diagrammatic representations of ADPs and their compositions.

\bigskip
\bibliographystyle{plain}
\bibliography{../../assets/references}

\end{document}